\documentclass{article}
\usepackage{spconf,amsmath,graphicx}

\usepackage{enumitem}
\usepackage{hyperref}

\hypersetup{hypertex=true,
            colorlinks=true,
            linkcolor=black,
            anchorcolor=black,
            citecolor=black}
            
\setlist{nosep, leftmargin=14pt}

\usepackage{mwe} 

\title{FDNet: Feature Decoupled Segmentation Network for Tooth CBCT Image}
\name{Xiang Feng$^{a,*}$,Chengkai Wang$^{a,*}$,Chengyu Wu$^{c}$,Yunxiang Li$^{d}$,Yongbo He$^{a}$,Shuai Wang$^{a}$,Yaqi Wang$^{b,\dagger }$
\thanks{ \begin{tabular}[t]{@{}l@{}} $^{*}$These authors contribute equally to this work.\\ $\dagger$ Corresponding author.\\
\end{tabular}}
       }
       
\address{$^a$ Hangzhou Dianzi University, Hangzhou, China\\
        $^b$ College of Media Engineering, Communication University of Zhejiang, Hangzhou, China\\
        $^c$ Department of Mechanical, Electrical and Information Engineering,
Shandong University, Weihai, China\\
        $^d$ University of Texas Southwestern Medical Center, Dallas, USA\\
        }

\usepackage{graphicx}
\usepackage{bbding}

\begin{document}
%
\maketitle
\begin{abstract}
Precise Tooth Cone Beam Computed Tomography (CBCT) image segmentation is crucial for orthodontic treatment planning. In this paper, we propose FDNet, a Feature Decoupled Segmentation Network, to excel in the face of the variable dental conditions encountered in CBCT scans, such as complex artifacts and indistinct tooth boundaries. The Low-Frequency Wavelet Transform (LF-Wavelet) is employed to enrich the semantic content by emphasizing the global structural integrity of the teeth, while the SAM encoder is leveraged to refine the boundary delineation, thus improving the contrast between adjacent dental structures. By integrating these dual aspects, FDNet adeptly addresses the semantic gap, providing a detailed and accurate segmentation. The framework’s effectiveness is validated through rigorous benchmarks, achieving the top Dice and IoU scores of 85.28\% and 75.23\%, respectively. This innovative decoupling of semantic and boundary features capitalizes on the unique strengths of each element to elevate the quality of segmentation performance.

\end{abstract}
\section{Introduction}
\label{sec:intro}

Tooth segmentation from Cone Beam Computed Tomography (CBCT) scans forms the crux of dental diagnostics, orthodontic treatment planning, and dental restoration procedures undertaken by stomatologists and dentists. The creation of precise tooth models depends on accurate segmentation, which has been traditionally a manual and tedious endeavor. The prevalence of low-quality images in CBCT scans, characterized by complex artifacts and blurred boundaries, complicates the task of segmentation, often leading to unclear demarcations and challenges in extracting accurate semantic information. The need for a robust, automated segmentation system capable of effectively managing common image quality problems and accurately outlining teeth is clearly evident.

Existing segmentation methodologies, notably U-Net \cite{unet} and its derivatives \cite{FCN,gt-UNet}, have shown effectiveness across various medical imaging domains. However, when applied to CBCT tooth segmentation, especially in 2D scenarios, a significant semantic gap between the encoder and decoder stages emerges, detrimentally impacting the segmentation performance~\cite{multiresunet, pang2019}. Furthermore, the inadequate fusion of semantic and boundary information, essential for precisely distinguishing adjacent dental structures, compounds the challenge\ cite{mea}. This highlights the pressing need for a bespoke solution designed to navigate the distinct hurdles associated with CBCT tooth imaging.

Addressing the outlined challenges, we introduce FDNet, a pioneering triple-branch network that effectively decouples and processes semantic and boundary features to overcome the challenges of CBCT tooth imaging. At its core are the LF-Wavelet and the SAM encoder  \cite{SAM}, tailored for semantic enrichment and boundary delineation in images, respectively. LF-Wavelet operates within the Low-frequency and Image Fusion (LIF) module, focusing on semantic information amplification, while the SAM encoder, a pre-trained Vision Transformer, excels at boundary feature extraction. Their synergistic operation mitigates the semantic gap between encoder and decoder stages and enhances resilience against common image defects. Through FDNet, we aim to mend the semantic gap and bolster the framework against typical image defects, advancing segmentation accuracy in CBCT tooth imaging.

The core contributions of our study are as follows:
\begin{itemize}
  \item We introduce FDNet, a distinctive triple-branch network that improves segmentation outcome by effectively decoupling and integrating semantic and boundary information.
  \item FDNet utilizes LIF to lessen the semantic gap between the encoder and decoder stages, which is a common drawback in preceding models.
  \item FDNet achieves superior performance on DICE and IoU over existing state-of-the-art models, providing a potential new baseline for tooth CBCT image segmentation.
\end{itemize}

\begin{figure*}[htbp]
\begin{center}

\includegraphics[width=0.97\textwidth]{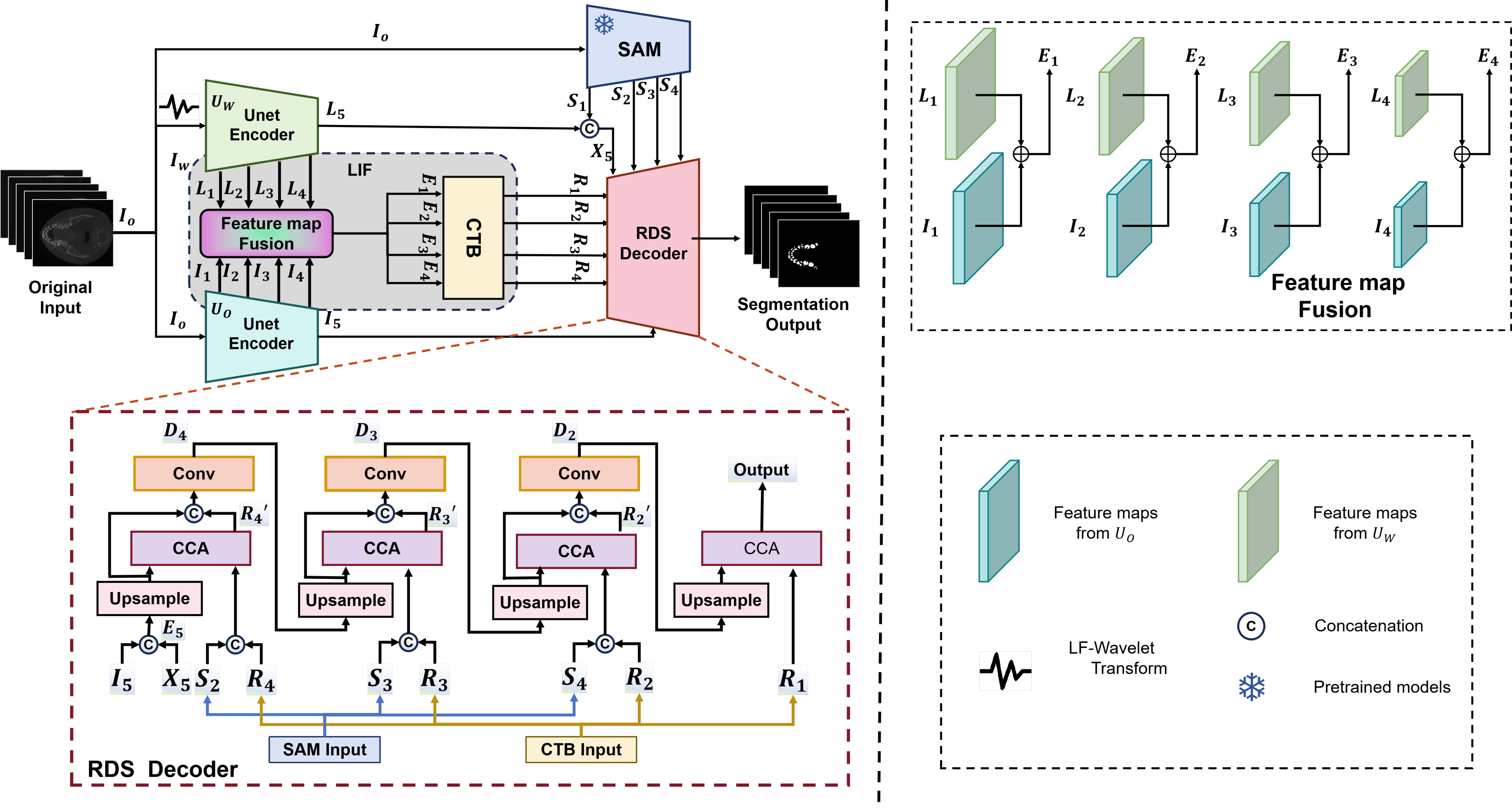}

\end{center}

\caption{The architecture of the proposed FDNet segmentation model. Specifically, the LIF module fuses features from the $U_w$ and $U_o$ encoders before passing them to the CTB module.}

\label{overview}

\end{figure*}

\section{METHODS}
\label{sec:format}

\subsection{Architecture Overview}
\label{ssec:subhead}

Our FDNet is a triple-branch model engineered for precision in tooth CBCT image segmentation, as shown in Fig.~\ref{overview}, epitomizes our innovative approach to feature decoupling, specifically targeting the nuanced intricacies of tooth CBCT images. It harnesses the LF-Wavelet and original image data, channeling these into the encoder before proceeding through the LIF’s Channel-wise Cross-attention Transformer Block (CTB) module for enhanced feature integration, capturing varied scale information effectively $(R_j, j \in\{1,2,3,4\})$. Subsequently, the refined features are seamlessly combined with the SAM encoder’s output, feeding into the RDS decoder to accomplish superior segmentation accuracy and boundary clarity.

\subsection{LF-Wavelet Transform}
\label{ssec:subhead}

\begin{figure}[htbp]
\centering
\hspace{-10mm}
\includegraphics[width=0.4\textwidth]{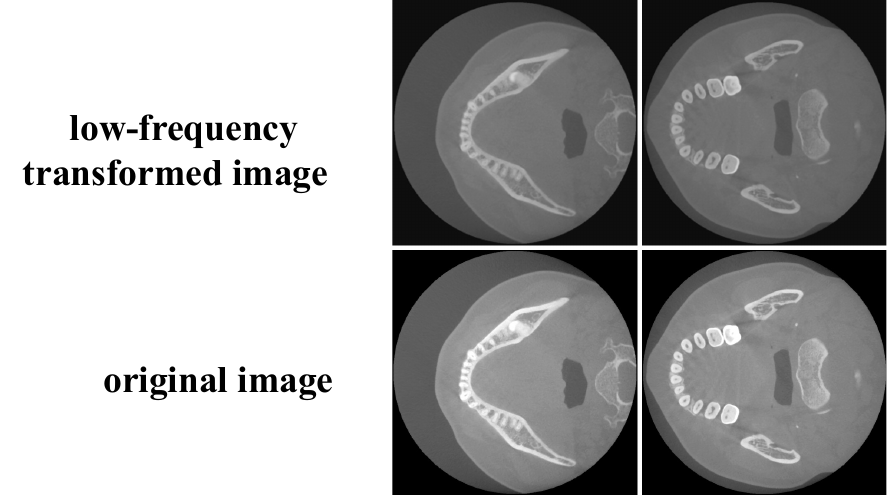}

\caption{Comparison of low-frequency transformed and original images, with each low-frequency counterpart on the top and its original image at the bottom.}
\label{compare}

\end{figure}

In the initial phase of our FDNet framework, we adopt an LF-Wavelet to address the semantic segmentation needs in CBCT tooth imaging. Accurate segmentation hinges on the inclusion of low-frequency (LF) semantics, such as overall shape and color consistency, which are crucial for understanding the global context of dental structures. To remedy this, we apply a wavelet decomposition to isolate and fuse the low-frequency details back into the original image, enriching the data fed into our deep-learning model. This process imbues the model with a fused representation of intricate semantic and structural information. Figure~\ref{compare} displays this process, allowing for a visual appreciation of how LF-Wavelet Transform prioritizes.

\begin{figure*}[htbp]
\begin{center}

\includegraphics[width=1\textwidth]{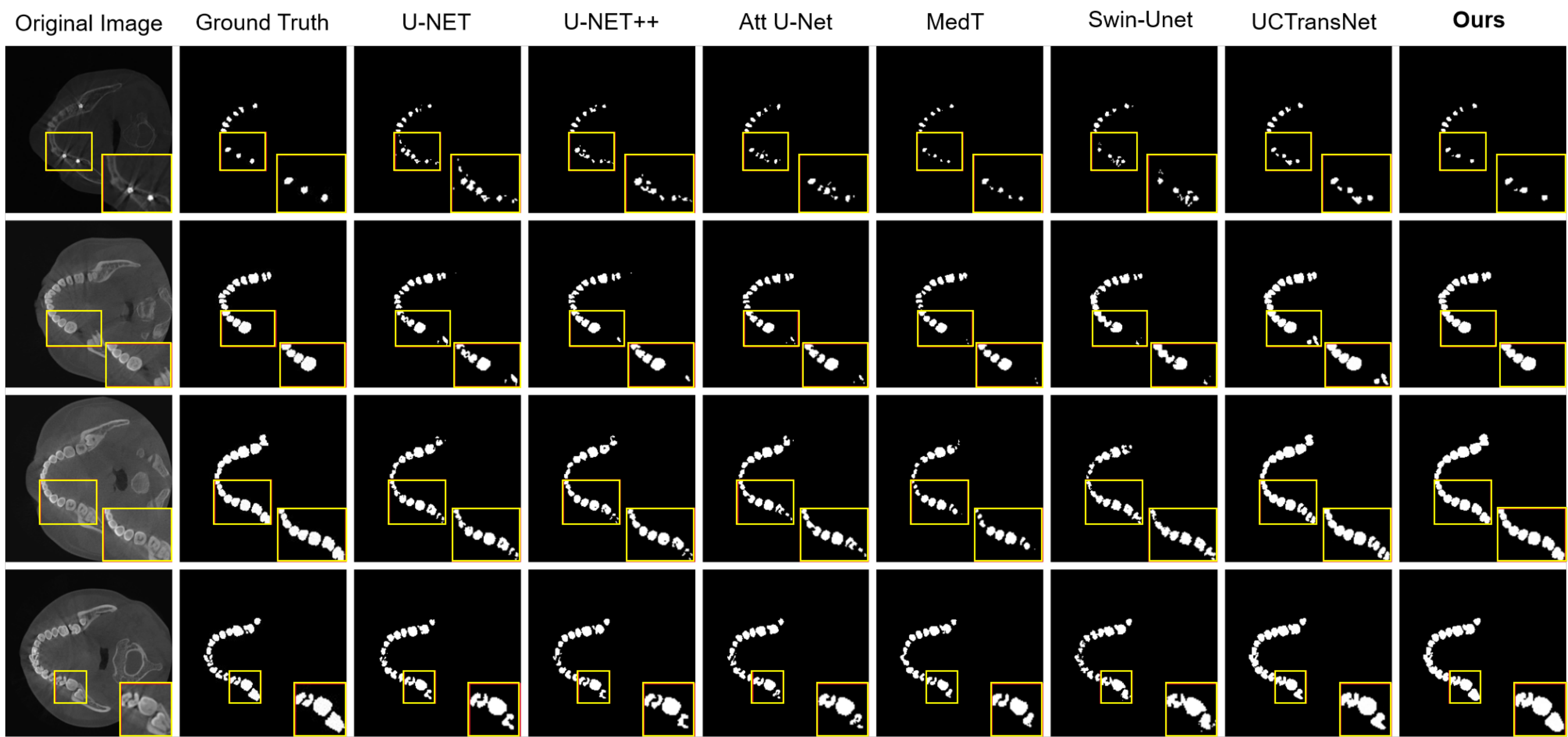}

\end{center}

\caption{The results of segmentation using different methods. It shows our model's adept performance in achieving precise segmentation across four dental conditions, including scenarios with complex artifacts and blurred tooth boundaries.}

\label{visual}

\end{figure*}
The mathematical expression for this LF enhancement is articulated as follows:
\begin{align}
   X_i^{\prime}=W^{-1}\left(L F\left(W\left(X_i\right)\right)\right).   \label{eq:LW}
\end{align}

Here, $X_i$ is the original input image, $W$ represents the wavelet transform, $W^{-1}$ is its inverse, and $L F$ is the extraction of low-frequency components. The enhanced image $X_i^{\prime}$ is thus a composite, fortified with low-frequency information, enhancing the segmentation process. By concatenating these low-frequency embeddings with the original image data, we construct a composite input that capitalizes on the strengths of both spectral domains in the first stage. Through this methodology, we mitigate the potential loss of contextual semantics and augment the model's proficiency in translating global features into precise local patterns, which is pivotal for refined and robust segmentation outcomes.

Moreover, the uniformity offered by the low-frequency details helps address segmentation challenges, such as small data interruptions - gaps or holes caused by high-frequency noise. Without this focus on low frequencies, such disruption could lead to fragmented segmentation, adversely affecting dental diagnostics and treatment planning.

\subsection{SAM Encoder}
\label{ssec:subhead}

After the low-frequency wavelet transformation, our FDNet framework incorporates the SAM Encoder to enhance feature representation for precise CBCT tooth segmentation. Pre-trained on the vast SA-1B segmentation dataset, the SAM Encoder is adept at general feature extraction, a capability essential for diverse imaging conditions.

The SAM Encoder's edge lies in its ability to highlight boundary features, which is critical for the clear delineation of adjacent dental structures within CBCT images. While the SAM Encoder is a powerful tool for feature extraction, its usage is optimized when guided by specific prompts, as it does not inherently understand the segmentation goals or the semantics involved.

We address this by employing a channel-wise cross attention (CCA) module \cite{UCTransNet} that synergistically combines the SAM Encoder's boundary-focused embeddings with the feature maps $R_j$ from the LIF module. The revised formula that depicts this integration is:
\begin{flalign}
& \left\{
\begin{aligned}
R_j' &= CCA(\text{Concat}(SAM(X_i), R_j), D_{j+1}) ,\\
D_j' &= \text{Conv}(\text{Concat}(R_j, D_{j+1})).
\end{aligned}
\right. & j &= \{2,3\} \\
& \left\{
\begin{aligned}
R_4' &= CCA(\text{Concat}(SAM(X_i), R_4), E_5), \\
E_5 &= \text{Concat}(SAM(X_i), L_5).
\end{aligned}
\right. &
\end{flalign}

Here, the enhanced feature map $R_j{}^{\prime}$ is produced by the CCA module, which integrates the boundary embeddings of the SAM Encoder with the output features from the CTB module and decoder features {$D_j,  j \in\{2,3\}$}.  For $R_4{}^{\prime}$, a unique process is employed: $S_2$ and $R_4$ are concatenated, then combined with $I_5$ and $X_5$ to produce $E_5$, which is subsequently processed by the CCA. Notably, $X_5$ itself is derived from concatenating the SAM-processed image result with the output $L_5$ from preceding UNet encoder operations.

This sophisticated concatenation and integration process ensures that each $R_j{}^{\prime}$ map is a comprehensive representation of both global and local contextual details, crucial for accurate dental anatomy segmentation. By fusing the SAM Encoder's boundary embeddings with the deep semantic and structural insights provided by the LF-Wavelet and the decoder feature map, our FDNet generates a feature map rich in detail and context. This integrated feature landscape empowers the RDS module, serving as the decoder, to effectively discern and segment complex dental structures in CBCT images with enhanced precision.

\section{EXPERIMENT}
\label{sec:pagestyle}

\subsection{Dataset}
\label{ssec:subhead}

Given the advancements in three-dimensional dental CBCT technology and the escalating importance of dentistry driven by economic upturn and an aging population, the task of tooth segmentation has gained prominence. In light of this, we leveraged a dataset provided by Hangzhou Dental Hospital, encompassing 9,000 labeled CBCT images, to accomplish our segmentation objectives. From this dataset, a subset of 288 images was randomly selected to constitute our training set, while a distinct set of 400 images was earmarked for testing purposes.

\begin{table}[h]
\caption{Evaluation comparison on multiple tooth segmentation models.}\label{tbl1}
\centering
\renewcommand{\arraystretch}{1.5} 

\resizebox{\linewidth}{!}{
\begin{tabular}{c|cccc}
\hline
& Dice (\%) & IoU (\%) & Recall & Precision\\
\hline
U-Net \cite{unet} & $81.69 \pm 10.66$ & $69.45 \pm 13.32$ & $82.14 \pm 10.06$ & $82.20 \pm 14.20$ \\
U-Net++ \cite{unet++} & $81.44 \pm 10.75$ & $69.93 \pm 13.69$ & $82.30 \pm 10.41$ & $82.47 \pm 14.31$ \\
Attention U-Net \cite{attUNet} & $81.45 \pm 10.66$ & $69.91 \pm 13.58$ & $83.10 \pm 9.49$ & $81.79 \pm 14.78$ \\
MedT \cite{MeDT} & $72.37 \pm 15.97$ & $58.85 \pm 17.20$ & $61.66 \pm 17.98$ & $\mathbf{92.19 \pm 13.51}$ \\
Swin-Unet \cite{swin-unet} & $75.23 \pm 14.51$ & $62.19 \pm 16.43$ & $75.90 \pm 16.39 $ & $77.44 \pm 14.59$ \\
UCTransNet \cite{UCTransNet} & $\underline{83.46 \pm 9.60}$ & $\underline{73.88 \pm 13.21}$ & $\mathbf{90.15 \pm 10.86}$ & $81.02 \pm 13.03$ \\

\textbf{FDNet (Ours)} & $\mathbf{85.28 \pm 9.78}$ & $\mathbf{75.23 \pm 13.08}$ & $\underline{89.77 \pm 10.99}$ & $\underline{82.67 \pm 12.47}$ \\
\hline
\end{tabular}
}
\label{evaluation}
\end{table}

\subsection{Main Results}
\label{ssec:subhead}

To evaluate the performance of our proposed FDNet framework, we conducted a comparative analysis against several advanced and robust models, including U-Net \cite{unet}, U-Net++ \cite{unet++}, Attention U-Net \cite{attUNet}, MedT \cite{MeDT}, Swin-Uet \cite{swin-unet}, and UCTransNet \cite{UCTransNet} on our tooth CBCT image dataset. All models were meticulously trained and tested based on their official source code and preset parameters.  We used a test set comprising 400 patient tooth samples.  The comparative visual results are illustrated in Fig.~\ref{visual}, while the quantitative evaluation is in Table~\ref{evaluation}. 

To highlight our model’s consistent performance across various patient cases and imaging conditions, we computed the average and standard deviation from multiple experiments. Our analysis revealed that FDNet achieved superior segmentation accuracy, surpassing the second-ranked model, UCTransNet, in both Dice and IoU metrics—indicators of segmentation correctness and the overlap between predicted and actual segmentation, respectively. While FDNet's average recall is marginally lower than that of UCTransNet, it showcases superior precision, with only MedT having a slightly higher value. However, it is essential to note that MedT's higher precision is countered by its lower recall, suggesting a trade-off between these metrics. Our model's balanced performance in both metrics means it maintains a harmonious balance between identifying relevant tooth structures (recall) and minimizing false positives (precision). This balance is critical in dental CBCT segmentation, where both over-segmentation and under-segmentation can have detrimental implications for subsequent dental treatment planning.

\subsection{Ablation Study}
\label{ssec:subhead}

An ablation study was orchestrated to dissect the individual contributions of our core modules: SAM encoder, LIF and LW-Wavelet, in the realm of tooth segmentation. As illustrated in Table~\ref{ablation}, this study illuminated the pivotal roles of each module in augmenting the segmentation quality. It underscored that the confluence of semantic information through LW-Wavelet, the robust feature extraction by the SAM encoder, and the fusion functionality of the LIF module synergistically drive superior segmentation outcomes.

\begin{table}[h]
\caption{Ablation Study Outlining the Impact of SAM Encoder, LIF, and LF-Wavelet on Segmentation Performance.}\label{tbl1}
\centering
\resizebox{\linewidth}{!}{%
\begin{tabular}{c|cccc}
\hline
Methods & Dice (\%) & IoU (\%) & Recall & Precision\\
\hline
FDNet (w/o SAM) & 85.24 & 75.14 & 87.56 & $\boldsymbol{84.77}$ \\
FDNet (w/o LIF) & 83.94 & 74.01 & 87.43 & 82.31 \\
FDNet (w/o LW) & 84.33 & 74.12 & 86.47 & 83.98 \\
\textbf{FDNet (Ours)} & $\boldsymbol{85.64}$ & $\boldsymbol{75.72}$ & $\boldsymbol{88.86}$ & $84.38$ \\
\hline

\end{tabular}%
}
\label{ablation}
\end{table}

\section{CONCLUSION}
\label{sec:typestyle}

In this work, we introduce FDNet, an innovative approach for tooth segmentation in CBCT imaging, addressing the inherent challenges presented by diverse dental conditions, indistinct tooth boundaries, and semantic paucity in CBCT images. FDNet's architectural ingenuity lies in its unique feature decoupling strategy. By incorporating the Low-Frequency Wavelet Transform within our framework, we enhance the semantic richness by capturing the global structural details from low-frequency components. Concurrently, the SAM encoder extracts boundary features critical for distinguishing adjacent dental structures. This strategic fusion of components facilitates a semantic bridge between the encoding and decoding phases, markedly elevating the segmentation accuracy, particularly in the face of complex dental scenarios. Our extensive testing underscores FDNet's superior performance over leading-edge models, highlighting its capacity to revolutionize automated tooth segmentation across various patient dental conditions and positioning it as a versatile and reliable framework in the realm of dental diagnostics and orthodontic treatment planning.

\section{COMPLIANCE WITH ETHICAL STANDARDS}
\label{sec:ethical standard}

This study was performed in line with the principles of the Declaration of Helsinki. Approval was granted by the Ethics Committee of Lishui University Medical School (N0.2022YR014).

\section{ACKNOWLEDGMENTS}
\label{sec:Acknowledgements}

 This research is supported in part by the National Natural Science Foundation of China [(No.62206242) and (No.62201323)].

\bibliographystyle{IEEEbib}
\bibliography{strings,refs}

\end{document}